\pdfoutput=1
\documentclass[11pt]{article}

\usepackage[final]{EMNLP2023}

\usepackage{times}
\usepackage{latexsym}

\usepackage[T1]{fontenc}
\usepackage[utf8]{inputenc}

\usepackage{microtype}

\usepackage{inconsolata}

\usepackage[utf8]{inputenc} % allow utf-8 input
\usepackage[T1]{fontenc}    % use 8-bit T1 fonts
\usepackage{hyperref}       % hyperlinks
\usepackage{url}            % simple URL typesetting
\usepackage{booktabs}       % professional-quality tables
\usepackage{amsfonts}       % blackboard math symbols
\usepackage{nicefrac}       % compact symbols for 1/2, etc.
\usepackage{microtype}      % microtypography
\usepackage{natbib}
\usepackage{adjustbox}
\graphicspath{ {./images/} }
\usepackage{amsmath}
\usepackage{makecell}

\usepackage{listings}
\usepackage{xcolor}

\usepackage{dblfloatfix}
\usepackage{placeins}

\usepackage{enumitem}
\usepackage{siunitx}

\definecolor{codegreen}{rgb}{0,0.6,0}
\definecolor{codegray}{rgb}{0.5,0.5,0.5}
\definecolor{codepurple}{rgb}{0.58,0,0.82}
\definecolor{backcolour}{rgb}{0.95,0.95,0.92}

\lstdefinestyle{mystyle}{
    backgroundcolor=\color{backcolour},   
    commentstyle=\color{codegreen},
    keywordstyle=\color{magenta},
    numberstyle=\tiny\color{codegray},
    stringstyle=\color{codepurple},
    basicstyle=\ttfamily\footnotesize,
    breakatwhitespace=false,         
    breaklines=true,                 
    captionpos=b,                    
    keepspaces=true,                 
    numbers=left,                    
    numbersep=5pt,                  
    showspaces=false,                
    showstringspaces=false,
    showtabs=false,                  
    tabsize=2
}

\title{EELBERT: Tiny Models through Dynamic Embeddings}

\author{Gabrielle Cohn, \  Rishika Agarwal, \  Deepanshu Gupta, \and Siddharth Patwardhan\\
  Apple\\
  Cupertino, CA 95014\\
  \texttt{\{gcohn,rishika\_agarwal,dkg,patwardhan.s\}@apple.com}
}
\begin{document}
\maketitle

\begin{abstract}

We introduce EELBERT, an approach for compression of transformer-based models (e.g., BERT), with minimal impact on the accuracy of downstream tasks.
This is achieved by replacing the input embedding layer of the model with dynamic, i.e. on-the-fly, embedding computations.
Since the input embedding layer accounts for a significant fraction of the model size, especially for the smaller BERT variants, replacing this layer with an embedding computation function helps us reduce the model size significantly.
Empirical evaluation on the GLUE benchmark shows that our BERT variants (EELBERT) suffer minimal regression compared to the traditional BERT models.
Through this approach, we are able to develop our smallest model UNO-EELBERT, which achieves a GLUE score within 4\% of fully trained BERT-tiny, while being 15x smaller (1.2 MB) in size.

\end{abstract}

\section{Introduction}
It has been standard practice for the past several years for natural language understanding systems to be built upon powerful pre-trained language models, such as BERT \cite{devlin2018bert}, T5 \cite{raffel2020exploring}, mT5 \cite{xue2021mt5}, and RoBERTa \cite{liu2019roberta}. These language models are comprised of a series of transformer-based layers, each transforming the representation at its input into a new representation at its output.
Such transformers act as the ``backbone'' for solving several natural language tasks, like text classification, sequence labeling, and text generation, and are primarily used to map (or {\em encode}) natural language text into a multidimensional vector space representing the semantics of that language.

Experiments in prior work \cite{kaplan2020scaling} have demonstrated that the size of the language model (i.e., the number of parameters) has a direct impact on task performance, and that 
increasing a language model's size improves its language understanding capabilities.
Most of the recent state-of-art results in NLP tasks have been obtained with very large models.
At the same time as massive language models are gaining popularity, however, there has been a parallel push to create much smaller models, which could be deployed in resource-constrained environments such as smart phones or watches.

Some key questions that arise when considering such environments:
{\em How does one leverage the power of such large language models on these low-power devices?}
{\em Is it possible to get the benefits of large language models without the massive disk, memory and compute requirements?} 
Much recent work in the areas of model pruning \cite{gordon2020compressing}, quantization \cite{Zafrir2019quantization}, distillation \cite{jiao2020tinybert, sanh2020distilbert} and more targeted approaches like the {\em lottery ticket hypothesis} \cite{chen2020lottery} aim to produce smaller yet effective models.
Our work takes a different approach by reclaiming resources required for representing the model's large vocabulary.

The inspiration for our work comes from \citet{ravi2018self}, who introduced 
dynamic embeddings, i.e. embeddings computed on-the-fly via hash functions.
We extend the usage of 
dynamic embeddings to transformer-based language models. 
We observe that 21\% of the trainable parameters in BERT-base \cite{turc2019} are in the embedding lookup layer. By replacing this input embedding layer with embeddings computed at run-time, we can reduce model size by the same percentage.

In this paper, we introduce an ``\underline{e}mb\underline{e}dding\underline{l}ess'' model -- EELBERT --  that uses a dynamic embedding computation strategy to achieve a smaller size.
We conduct a set of experiments to empirically assess the quality of these ``embeddingless'' models along with the relative size reduction. 
A size reduction of up to 88\%  is observed in our experiments, with minimal regression in model quality, and this approach is entirely complementary to other model compression techniques.
Since EELBERT calculates embeddings at run-time, we do incur additional latency, which we measure in our experiments.
We find that EELBERT's latency increases relative to BERT's as model size decreases, but could be mitigated through careful 
architectural and engineering optimizations.
Considering the gains in model compression that EELBERT provides, this is not an unreasonable trade-off.

\section{Related Work}

There is a large body of work describing strategies for optimizing memory and performance of the BERT models \cite{Ganesh_2021}.
In this section, we highlight the studies most revelant to our work, which focus on reducing the size of the token embeddings used to map input tokens to a real valued vector representation.
We also look at past research on hash embeddings or randomized embeddings used in language applications (e.g., \citet{tito2017hash}).

Much prior work has been done to reduce the size of pre-trained static embeddings like GloVe and Word2Vec.
\citet{lebret-collobert-2014-word} apply Principal Component Analysis (PCA) to reduce the dimensionality of word embedding. 
For compressing GloVe embeddings, \citet{arora-etal-2018-linear} proposed LASPE, which leverages matrix factorization to represent the original embeddings as a combination of basis embeddings and linear transformations. 
\citet{lam2018word2bits} proposed a method called Word2Bits that uses quantization to compress Word2Vec embeddings. 
Similarly, \citet{kim-etal-2020-adaptive} proposed using variable size code-blocks to represent each word, where the codes are learned via a feedforward network with binary constraint. 

However, the most relevant works to this paper are by \citet{ravi-kozareva-2018-self} and \citet{ravi2017projectionnet}. 
The key idea in the approach by \citet{ravi-kozareva-2018-self} is the use of projection networks as a deterministic function to generate an embedding vector from a string of text, where this generator function replaces the embedding layer.

That idea has been extended to word-level embeddings by \citet{sankar2021proformer} and \citet{ravi-kozareva-2021-soda}, using an LSH-based technique for the projection function. 
These papers demonstrate the effectiveness of projection embeddings, combined with a stacked layer of CNN, BiLSTM and CRF, on a small text classification task. 
In our work, we investigate the potential of these projection and hash embedding methods to achieve compression in transformer models like BERT.

\begin{figure}[t]
    \centering
    \includegraphics[scale=.25]{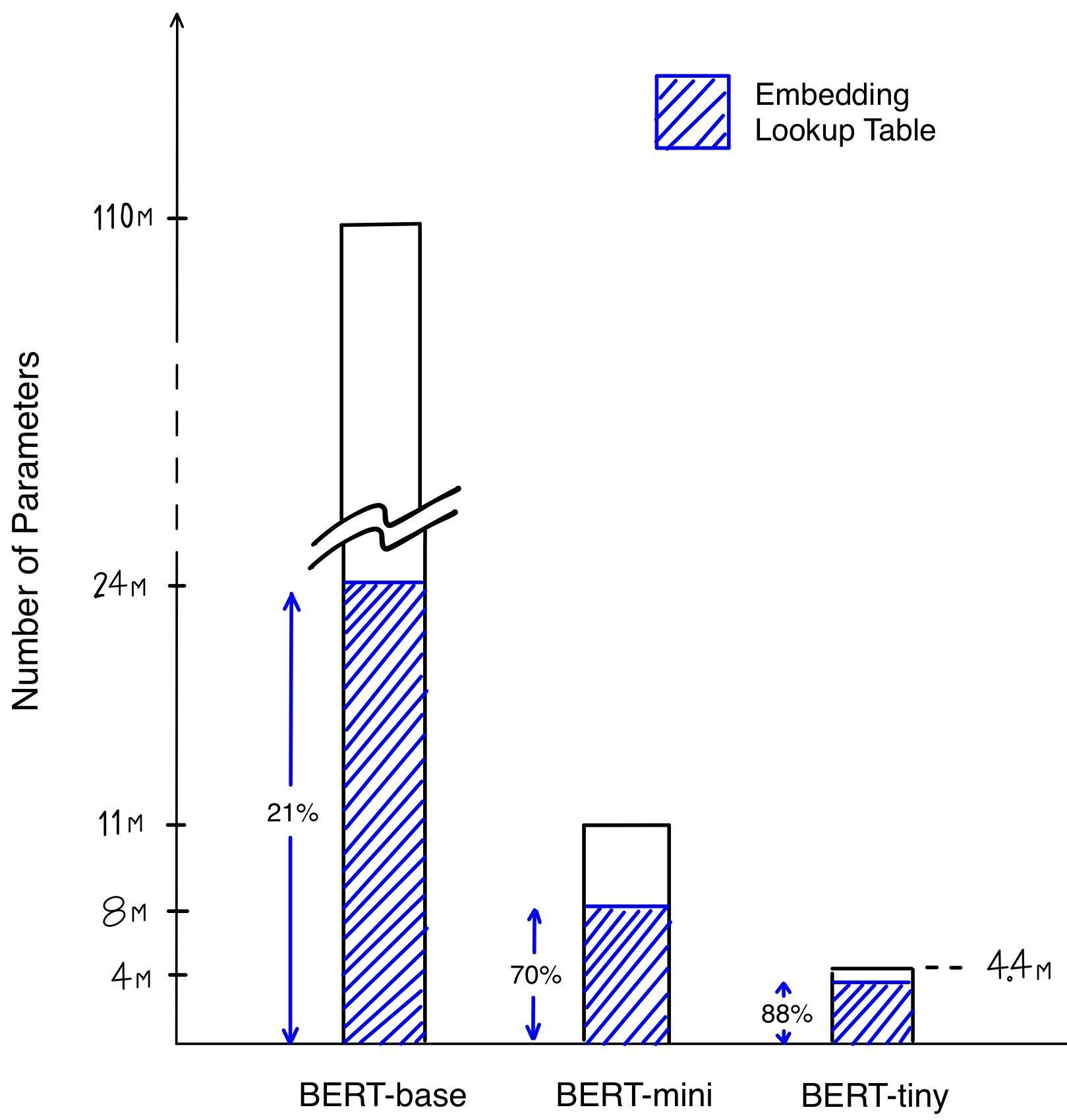}
    \caption{Embedding table in BERT}
    \label{fig:embtable}
\end{figure}

\section{Modeling EELBERT}
EELBERT is designed with the goal of reducing the size (and thus the memory requirement) of 
the input embedding layers of BERT and other transformer-based models.
In this section, we first describe our observations about BERT which inform our architecture choices in EELBERT, and then present the EELBERT model in detail.

\subsection{Observations about BERT}
BERT-like language models take a sequence of tokens as input, encoding them into a semantic vector space representation.
The input tokens are generated by a tokenizer, which segments a natural language sentence into discrete sub-string units $w_1, w_2, \ldots, w_n$.
In BERT, each token in the model's vocabulary is mapped to an index, corresponding to a row in the input embedding table (also referred to as the input embedding layer). This row represents the token's $d$-size embedding vector $\mathbf{e_{w_i}} \in \mathbb{R}^d$, for a given token $w_i$.

The table-lookup-like process of mapping tokens in the vocabulary to numerical vector representations using the input embedding layer
is a ``non-trainable'' operation, and is therefore unaffected by standard model compression techniques, which typically target the model's trainable parameters. This results in a compression bottleneck, since 
a profiling of BERT-like models reveals that the input embedding layer occupies a large portion of the model's parameters.

We consider three publicly available BERT models of different sizes, all pre-trained for English \cite{turc2019} -- {\em BERT-base}, {\em BERT-mini} and {\em BERT-tiny}.
BERT-base has 12 layers with a hidden layer size of 768, resulting in about 110M trainable parameters. BERT-mini has 4 layers and a hidden layer size of 256, with around 11M parameters, and 
BERT-tiny has 2 layers and a hidden layer size of 128, totaling about 4.4M parameters.

Figure~\ref{fig:embtable} shows the proportion of model size occupied by the input embedding layer
(blue shaded portion of the bars) versus the encoder
layers (un-shaded portion of the bars).
Note that in the smallest of these BERT variants, BERT-tiny, the input embedding layer occupies almost 90\% of the model.
By taking a different approach to model compression, focusing not on reducing the trainable parameters but instead on
eliminating the input embedding layer, one could potentially deliver up to 9x model size reduction.

\subsection{EELBERT Architecture}

EELBERT differs from BERT only in the process of going from input token to input embedding. 
Rather than looking up each input token in the input embedding layer as our first step, we dynamically compute an embedding for a token $w_i$ by using an $n$-gram pooling hash function.
The output is a $d$-size vector representation, $\mathbf{e_{w_i}} \in \mathbb{R}^d$, just as we would get from the embedding layer in standard BERT.
Keep in mind that EELBERT only impacts token embeddings, not the segment or position embeddings, and that all mentions of ``embeddings'' hereafter refer to token embeddings.

The key aspect of this method
is that it does not rely on an input embedding table stored in memory,
instead using the hash function to map input tokens to embedding vectors at runtime.
This technique is not intended to produce embeddings that approximate BERT embeddings.
Unlike BERT's input embeddings, dynamic embeddings do not update during training.

\begin{figure}[t]
    \centering
    \includegraphics[scale=.10]{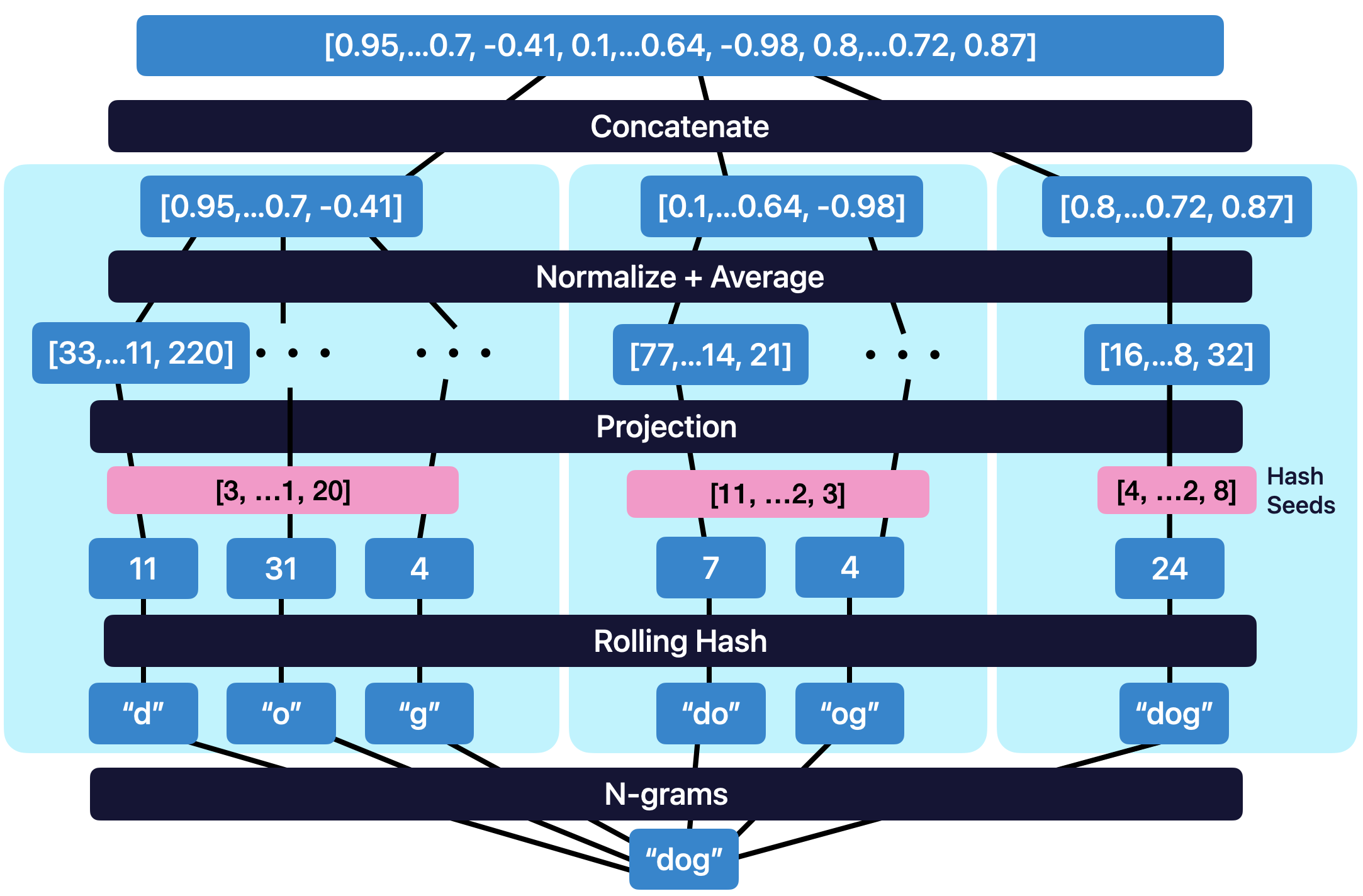}
    \caption{Computing dynamic hash embeddings}
    \label{fig:hash-steps}
\end{figure}

Our $n$-gram pooling hash function methodology is shown in Figure~\ref{fig:hash-steps}, with operations in black boxes, and black lines going from the input to the output of those operations.
Input and output values are boxed in blue.
For ease of notation, we refer to the $n$-grams of length $i$ as $i$-grams, where $i=1,...,N$, and $N$ is the maximum $n$-gram size. The steps of the algorithm are as follows:

\noindent{\bf 1. Initialize random hash seeds $\mathbf{h} \in \mathbb{Z}^d$.}
There are $d$ hash seeds in total, where $d$ is the size of the embedding we wish to obtain, e.g. 768 for BERT-base. The $d$ hash seeds are generated via a fixed random state, so we only need to save a single integer specifying the random state.

\noindent{\bf 2. Hash $i$-grams to get $i$-gram signatures $\mathbf{s_i}$.} There are $k_i=l-i+1$ number of $i$-grams, where $l$ is the length of the token. Using a rolling hash function \cite{enwiki:1168768744}, we compute the $i$-gram signature vectors, $\mathbf{s_i} \in \mathbb{Z}^{k_i}$.

\noindent{\bf 3. Compute projection matrix for $i$-grams.}
For each $i$,
we compute a projection matrix $\mathbf{P_i}$ using a subset of the hash seeds. 
The hash seed vector $\mathbf{h}$ is partitioned into $N$ vectors, boxed in pink in the diagram. Each partition $\mathbf{h_i}$ is of length $d_i$, where $\sum_{i=1}^Nd_i = d$, with larger values of $i$ corresponding to a larger $d_i$. 
Given the hash seed vector $\mathbf{h_i}$ and the $i$-gram signature vector $\mathbf{s_i}$, the projection matrix $\mathbf{P_i} \in \mathbb{Z}^{k_i \times d_i}$
is the outer product $\mathbf{s_i} \times \mathbf{h_i}$. 
To ensure that the matrix values are bounded between $[-1, 1]$, we perform a sequence of transformations on $\mathbf{P_i}$: 
\begin{align*} 
\mathbf{P_i} &= \mathbf{P_i}\ \%\ B \\
\mathbf{P_i} &= \mathbf{P_i} - (\mathbf{P_i} >\frac{B}{2})*B \\
\mathbf{P_i} &= \mathbf{P_i}\ /\ \frac{B}{2}
\end{align*}
where $B$ is our bucket size (scalar).

\noindent{\bf 4. Compute embedding, $\mathbf{e_i}$, for each $i$-grams.}
We obtain $\mathbf{e_i} \in \mathbb{R}^{d_i}$ by averaging $\mathbf{P_i}$ across its $k_i$ rows to produce a single $d_i$-dimensional vector.

\noindent{\bf 5. Concatenate $\mathbf{e_i}$ to get token embedding $e$.}

We concatenate the $N$ vectors $\{\mathbf{e_i}\}_{i=1}^N$,
to get the token's final embedding vector, $\mathbf{e} \in \mathbb{R}^d$.

For a fixed embedding size $d$, the tunable hyper-parameters of this algorithm are: $N$, $B$,
and the choice of the hashing function. We used $N=3$, $B=10^9+7$  and rolling hash function.

Since EELBERT replaces the input embedding layer with dynamic embeddings, the exported model size is reduced by the size of the input embedding layer:
$O(d \times V)$
where $V$ is the vocabulary size, and $d$ is the embedding size.

We specifically refer to the {\em exported size} here, because during pre-training, the model also uses an output embedding layer which maps embedding vectors back into tokens.
In typical BERT pre-training, weights are shared between the input and output embedding layer, so the output embedding layer does not contribute to model size.
For EELBERT, however, there is no input embedding layer to share weights with, so the output embedding layer does contribute to model size.
Even if we pre-compute and store the dynamic token embeddings
as an embedding lookup table, using the transposed dynamic embeddings as a frozen output layer would
defeat the purpose of learning contextualized representations.
In short, using coupled input and output embedding layers in EELBERT is infeasible, so BERT and EELBERT are the same size during pre-training.
When pre-training is completed, the output embedding layer in both models is discarded, and the exported models are used for downstream tasks, which is when we see the size advantages of EELBERT.

\section{Experimental Setup}

In this section, we assess the effectiveness of EELBERT. The key questions that interest us are: {\em how much model compression can we achieve} and {\em what is the impact of such compression on model quality for language understanding?}
We conduct experiments on a set of benchmark NLP tasks to empirically answer these questions.

In each of our experiments, we compare EELBERT to the corresponding standard BERT model -- i.e., a model with the same configuration but with the standard trainable input embedding layer instead of our dynamic embeddings. This standard model serves as the baseline for comparison, to observe the impact of our approach.

\subsection{Pre-training}
\label{sec:pre-training}

For our experiments, we pre-train both BERT and EELBERT from scratch on the OpenWebText dataset \cite{radford2019language, Gokaslan2019OpenWeb}, using the pre-training pipeline released by Hugging Face Transformers \cite{wolf2020huggingfaces}.
Each of our models is pre-trained for 900,000 steps with a maximum token length of 128 using the {\em bert-base-uncased} tokenizer.
We follow the pre-training procedure described in \citet{devlin2018bert}, with a few differences.
Specifically, (a) we use the OpenWeb Corpus for pre-training, while the original work used the combined dataset of Wikipedia and BookCorpus,
and (b) we only use the {\em masked language model} pre-training objective, while the original work employed both {\em masked language model} and {\em next sentence prediction} objectives.

For BERT, the input and output embedding layers are coupled and trainable. Since EELBERT has no input embedding layer, its output embedding layer is decoupled and trainable.

\subsection{Fine-tuning}
\label{sec:fine-tuning}

For downstream fine-tuning and evaluation, we choose the GLUE benchmark \cite{wang2019glue} to assess the quality of our models. 
GLUE is a collection of nine language understanding tasks, 
including single sentence tasks (sentiment analysis, linguistic acceptability), similarity/paraphrase tasks, and natural language inference tasks.
Using each of our models as a backbone, we fine-tune individually for each of the GLUE tasks under a setting similar to that described in \citet{devlin2018bert}.
The metrics on these tasks serve as a proxy for the quality of the embedding models.
Since GLUE metrics are known to have high variance, we run each experiment 5 times using 5 different seeds, and report the median of the metrics on all the runs, as done in \citet{lan2020albert}.

\begin{table}
\centering
\footnotesize
\begin{tabular}{|r|c|c|c|}
\hline
\multicolumn{1}{|c|}{} & {\scriptsize \bf BERT-base} & {\scriptsize \bf EELBERT-base}\\
\hline
Trainable Parameters & 109,514,298 & {\bf 86,073,402}\\
Exported Model Size & 438 MB & {\bf 344 MB}\\
\hline
SST-2 (Acc.) & 0.899 & 0.900\\
QNLI (Acc.) & 0.866 & 0.864\\
RTE (Acc.) & 0.625 & 0.563\\
WNLI* (Acc.) & 0.521 & 0.563\\
MRPC (Acc., F1) & 0.833, 0.882 & 0.838, 0.887\\

QQP* (Acc., F1) & 0.898, 0.864 & 0.895, 0.861\\
MNLI (M, MM Acc.) & 0.799, 0.802 & 0.790, 0.795\\

STSB (P, S Corr.) & 0.870, 0.867 & 0.851, 0.849\\
CoLA (M Corr.) & 0.410 & 0.373\\

\hline

GLUE Score & 0.775 & 0.760\\
\hline
\end{tabular}
\caption{GLUE benchmark for BERT vs. EELBERT}
\label{tab:exp1}
\end{table}

\begin{table*}[t]
  \centering
  \footnotesize
  \begin{tabular}{|r|c|c|c|c|c|c|c|}
    \hline
    \multicolumn{1}{|c|}{} & {\bf BERT-mini} & {\bf EELBERT-mini} & {\bf BERT-tiny} & {\bf EELBERT-tiny} & {\bf UNO-EELBERT}\\
    \hline
    Trainable Parameters & 11,171,074 & {\bf 3,357,442} & 4,386,178 & {\bf 479,362} & {\bf 312,506}\\
    \hline
    Exported Model Size & 44.8 MB  &  {\bf 13.4 MB} & 17.7 MB & {\bf 2.04 MB} & {\bf 1.24 MB}\\
    \hline
    SST-2 (Acc.) &0.851 & 0.835 & 0.821 & 0.749 & 0.701\\

    QNLI (Acc.)  & 0.827 & 0.821 & 0.616 & 0.705 & 0.609 \\
    RTE (Acc.) & 0.552 & 0.560 &0.545 & 0.516 & 0.527\\
    WNLI* (Acc.) &0.563 & 0.549  &0.521 & 0.535 & 0.479 \\
    MRPC (Acc., F1) &  0.701, 0.814 & 0.721, 0.814   &0.684, 0.812 & 0.684, 0.812 & 0.684,0.812\\
    QQP* (Acc., F1) & 0.864, 0.815 & 0.850, 0.803 & 0.780, 0.661 & 0.752, 0.712 & 0.728, 0.628\\
    MNLI (M, M Acc.)  & 0.719, 0.730  & 0.688, 0.697 & 0.577, 0.581 & 0.582, 0.598 & 0.539, 0.552\\

    CoLA  (M Corr.)  &   0.103  & 0 & 0 & 0 & 0\\
    \hline
    GLUE score & 0.753 & 0.746 & 0.671 & 0.666 & 0.632 \\
    
    \hline 
  \end{tabular}
  \caption{EELBERT with smaller models}
  \label{tab:exp2}
\end{table*}

We calculate an overall GLUE score for each model. For BERT-base and EELBERT-base we use the following equation:

\begin{lstlisting}[numbers=none]
AVERAGE(CoLA Matthews corr, SST-2 accuracy, MRPC accuracy, STSB Pearson corr, QQP accuracy, AVERAGE(MNLI match accuracy, MNLI mismatch accuracy), QNLI accuracy, RTE accuracy)
\end{lstlisting}

Like \citet{devlin2018bert}, we do not include the WNLI task in our calculations.
For all the smaller BERT variants, i.e. BERT-mini, BERT-tiny, EELBERT-mini, EELBERT-tiny, and UNO-EELBERT, we use:

\begin{lstlisting}[numbers=none]
AVERAGE(SST-2 accuracy, MRPC accuracy, QQP accuracy, AVERAGE(MNLI match accuracy, MNLI mismatch accuracy), QNLI accuracy, RTE accuracy)
\end{lstlisting}

Note that we exclude CoLA and STSB from the smaller models' score, because the models (both baseline and EELBERT) appear to be unstable on these tasks. We see a similar exclusion of these tasks in \citet{sun2019patient}.

Also note that in the tables we abbreviate MNLI match and mismatch accuracy as MNLI (M, MM Acc.), CoLA Matthews correlation as CoLA (M Corr.), and STSB Pearson and Spearman correlation as STSB (P, S Corr.).

\begin{table*}[t]
  \centering
  \small
  \begin{tabular}{|r|c|c|c|c|c|c|c|}
    \hline
    \multicolumn{1}{|c|}{} & \multicolumn{2}{c|} {\bf BERT-base} & \multicolumn{2}{c|} {\bf BERT-mini}\\
    \hline
    Initialization Method & $n$-gram pooling & random & $n$-gram pooling & random\\
    \hline
    Trainable Parameters & 86,073,402 & 86,073,402 & 3,387,962 & 3,387,962 \\
    \hline
    Exported Model Size & 344 MB  &  344 MB & 13.4 MB & 13.4 MB \\
    \hline
   
    SST-2 (Acc.) & 0.900 & 0.897 & 0.835 & 0.823 \\
    QNLI (Acc.)  & 0.864 & 0.862 & 0.821 & 0.639  \\
    RTE (Acc.) & 0.563 & 0.574 & 0.560 & 0.569 \\
    WNLI* (Acc.) & 0.563 & 0.507  & 0.549 & 0.507  \\
    MRPC (Acc., F1) &  0.838, 0.887 & 0.806, 0.868 & 0.721, 0.814 & 0.690, 0.805 \\

    QQP* (Acc., F1) & 0.895, 0.861 & 0.893, 0.858 & 0.850, 0.803 & 0.800, 0.759\\
    MNLI (M, MM Acc.)  & 0.791, 0.795  & 0.786, 0.794 & 0.688, 0.697 & 0.647, 0.660\\
    STSB (P, S Corr.) & 0.851, 0.849 & 0.849, 0.847 &  -,- & -,- \\
    CoLA  (M Corr.)  &   0.373  & 0.389 & 0 & 0 \\
    \hline
    GLUE score & 0.760 & 0.757 & 0.746 & 0.696 \\
    
    \hline 
  \end{tabular}
  \caption{Impact of varying hash functions}
  \label{tab:exp3}
\end{table*}

\section{Results}

We present results of experiments assessing various aspects of the model with a view towards deployment and production use.

\subsection{Model Size vs. Quality}

Our first experiment directly assesses our dynamic embeddings by comparing the EELBERT models to their corresponding standard BERT baselines on GLUE benchmark tasks.
We start by pre-training the models as described in Section \ref{sec:pre-training}
and fine-tune the models on downstream GLUE tasks, as described in Section \ref{sec:fine-tuning}.

Table \ref{tab:exp1} summarizes the results of this experiment.
Note that replacing the trainable embedding layer with dynamic embeddings does have a relatively small impact on the GLUE score.
EELBERT-base achieves $\sim$21\% reduction in parameter count while regressing by just 
1.5\% on the GLUE score.

As a followup to this, we investigate the impact of dynamic embeddings on significantly smaller sized models.
Table \ref{tab:exp2} shows the results for BERT-mini and BERT-tiny, which have 11 million and 4.4 million trainable parameters, respectively.
The corresponding EELBERT-mini and EELBERT-tiny models have 3.4 million and 0.5 million trainable parameters, respectively. 
EELBERT-mini has just 0.7\% absolute regression compared to BERT-mini, while being $\sim$3x smaller. Similarly, EELBERT-tiny is almost on-par with BERT-tiny, with 0.5\% absolute regression, while being $\sim$9x smaller.

Additionally, when we compare EELBERT-mini and BERT-tiny models, which have roughly the same number of trainable parameters, we notice that EELBERT-mini has a substantially higher GLUE score than BERT-tiny. 
This leads us to conclude that under space-limited conditions, it would be better to train a model with dynamic embeddings and a larger number of hidden layers rather than a shallower model with trainable embedding layer and fewer hidden layers.

\subsection{Pushing the Limits: UNO-EELBERT}

The results discussed in the previous section suggest that our dynamic embeddings have the most utility for extremely small models, where they perform comparably
to standard BERT while providing drastic compression. 
Following this line of thought, we try to push the boundaries of model compression.
We train UNO-EELBERT, a model with a similar configuration as EELBERT-tiny, but a reduced intermediate size of 128.
We note that this model is almost 15 times smaller than BERT-tiny, with an absolute GLUE score regression of less than 
4\%.
It is also 350 times
smaller than BERT-base, with an absolute regression of less than 
20\%. Note that for these regression calculations, all GLUE scores were calculated using the small-model GLUE score equation, which excludes CoLA and STSB, so that the scores would be comparable.
We believe that with a model size of 1.2 MB, UNO-EELBERT could be a powerful candidate for low-memory edge devices like IoT, and other memory critical applications.

\subsection{Impact of Hash Function}

Our results thus far suggest that the trainable embedding layer can be replaced by a deterministic hash function with minimal impact on downstream quality.
The hash function we used pools the $n$-gram features of a word to generate its embedding, so words with similar morphology, like "running" and "runner", will result in similar embeddings.
In this experiment, we investigate whether our particular choice of hash function plays an important role in the model quality, or whether a completely random hash function which preserves no morphological information would yield similar results.

\begin{table}
\centering
\footnotesize
\begin{tabular}{|r|c|c|c|}
\hline
\multicolumn{1}{|c|}{} & \multicolumn{2}{c|}{\bf BERT-base }  \\ 
    \hline
    Initialization Method & random & hash  \\
    \hline
    Trainable Parameters & 109,514,298 & 109,514,298    \\
    \hline
    Exported Model Size & 438 MB & 438 MB\\
    \hline
    
    SST-2 (Acc.) & 0.899 & 0.904 \\
    QNLI (Acc.) & 0.866 & 0.876 \\
    RTE (Acc.) & 0.625 & 0.614  \\
    WNLI* (Acc.) & 0.521 & 0.563  \\
    MRPC (Acc., F1) & 0.833, 0.882 & 0.850, 0.896  \\
    
    QQP* (Acc., F1) & 0.898, 0.864 & 0.901, 0.867   \\
    MNLI (M, MM Acc.) & 0.799, 0.802 & 0.807, 0.809  \\
    
    STSB (P, S Corr.) & 0.870, 0.867 & 0.869, 0.867 \\
    
    CoLA  (M Corr.)  & 0.410 & 0.417  \\
    \hline
    GLUE score & 0.775 & 0.780   \\
    \hline 
\end{tabular}
\caption{Initialization of trainable embeddings}
\label{tab:exp4}
\end{table}

To simulate a random hash function, 
we initialize the embedding layer of BERT with a random normal distribution (BERT's default initialization scheme), and then freeze the embedding layer, so each word in the vocabulary is mapped to a random embedding.
The results presented in Table \ref{tab:exp3} indicate that for larger models like BERT-base, the hashing function doesn't have much significance, as the models trained with random vs $n$-gram pooling hash functions perform similarly on the GLUE tasks.
However, for the smaller BERT-mini model, our $n$-gram pooling hash function results in a better score.
These results suggest
that the importance of the $n$-gram pooling hash function, as compared to a completely random hash function, increases as the model size decreases.
This is a useful finding, since the primary benefit of dynamic hashing is to develop small models that can be run on device.

\begin{table*}[t]
  \centering
  \small
  \begin{tabular}{|r|c|c|c|c|c|c|c|}
    \hline
    \multicolumn{1}{|c|}{} & \multicolumn{1}{c|}{\scriptsize \bf BERT-base} & \multicolumn{1}{c|}{\scriptsize \bf EELBERT-base} & \multicolumn{1}{c|}{\scriptsize \bf BERT-mini} & \multicolumn{1}{c|}{\scriptsize \bf EELBERT-mini} & \multicolumn{1}{c|}{\scriptsize \bf BERT-tiny} & \multicolumn{1}{c|}{\scriptsize \bf EELBERT-tiny}\\
    \hline
    Model Size (MB) & 428.00 & 344.00 & 44.80 & 13.40 & 17.40 & 2.04 \\
    \hline
    Latency (ms) & 162.0 & 165.0 & 7.0 & 9.9 & 1.7 & 3.9 \\
    \hline
  \end{tabular}
  \caption{Latency, on MacBookPro M1 32GB RAM}
  \label{tab:latency}
\end{table*}

\subsection{Hash Function as Initializer}

Based on the results of the previous experiment, we consider a potential alternative role for the embeddings generated by our hash function.
We investigate whether our $n$-gram pooling hash function could be a better {\em initializer} for a trainable embedding layer, compared to the commonly used random normal distribution initializer.
To answer this question, we conduct an experiment with BERT-base, by intializing one model with the default random normal initialization and the other model with the embeddings generated using our $n$-gram pooling hash function ({\em hash} column in Table \ref{tab:exp4}).
Note that in this experiment the input and output embedding layers are coupled, and embedding layers are trainable for both initialization schemes.

The results of this experiment are shown in Table \ref{tab:exp4}.
The hash-initialized model shows a 0.5\% absolute increase in GLUE score compared to the randomly-initialized model.
We also perform this comparison for BERT-mini (not shown in the table), and observe a similar result.
In fact, for BERT-mini, the hash-initialized model had an absolute increase of
1.6\% 
in overall GLUE score, suggesting that the advantage of $n$-gram pooling hash-initialization may be even greater for smaller models.

\subsection{Memory vs. Latency Trade-off}
One consequence of using dynamic embeddings is that we are essentially trading off computation time for memory. 
The embedding lookup time for a token is $O(1)$ in BERT models.
In EELBERT, token embedding depends on the number of character $n$-grams in the token, as well as the size of the hash seed partitions.
Due to the outer product between the $n$-gram signatures and the partitioned hash seeds, the overall time complexity is dominated by $l \times d$, where $l$ is the length of a token, and $d$ is the embedding size, leading to $O(l \times d)$ time complexity to compute the dynamic hash embedding for a token. For English, the average number of letters in a word follows a somewhat Poisson distribution, with the mean being $\sim$4.79 \cite{Norvig2012},
and the embedding size $d$ for BERT models typically ranging between 128 to 768.

The inference time for BERT-base vs EELBERT-base is practically unchanged, as the bulk of the computation time goes in the encoder blocks for big models with multiple encoder blocks.
However, our experiments in Table \ref{tab:latency} indicate that EELBERT-tiny has $\sim$2.3x the inference time of BERT-tiny, as the computation time in the encoder blocks decreases for smaller models, and embedding computation starts constituting a sizeable portion of the overall latency.
These latency measurements were done on a standard M1 MacBook Pro with 32GB RAM. We performed inference on a set of 10 sentences (with average word length of 4.8) for each of the models, reporting the average latency of obtaining the embeddings for a sentence (tokenization latency is same for all the models, and is excluded from the measurements).

To improve the inference latency, we suggest some architectural and engineering optimizations.
The outer product between the $O(l)$ dimensional $n$-gram hash values and $O(d)$ dimensional hash seeds, resulting in a matrix of size
$O(l \times d)$, is the computational bottle-neck in the dynamic embedding computation.
A sparse mask with a fixed number of 1's in every row could reduce the complexity of this step to $O(l \times s)$, where $s$ is the number of ones in each row, and $s \ll d$.
This means every $n$-gram will only attend to some of the hash seeds.
This mask can be learned during training, and saved with the model parameters without much memory overhead, as it would be of size $O(k \times s)$, $k$ being the max number of $n$-grams expected from a token.
Future work could explore the effect of this approach on model quality.
The hash embedding of tokens could also be computed in parallel, since they are independent of each other.
Additionally, we observe that the 1, 2 and 3-grams follow a Zipf-ian distribution.
By using a small cache of the embeddings for the most common $n$-grams, we could speed up the computation at the cost of a small increase in memory footprint.

\section{Conclusions} 

In this work we explored the application of dynamic
embeddings to the BERT model architecture, as an alternative to the standard, trainable input embedding layer. Our experiments show that replacing the input embedding layer with 
dynamically computed embeddings is an effective method of model compression, with minimal regression on downstream tasks.
Dynamic embeddings appear to be particularly effective for the smaller BERT variants, where the input embedding layer comprises a larger percentage of trainable parameters.  

We also find that for smaller BERT models, a deeper model with dynamic embeddings yields better results than a shallower model of comparable size with a trainable embedding layer.
Since the dynamic embeddings technique used in EELBERT is complementary to existing model compression techniques, we can apply it in combination with other compression methods to produce extremely tiny models. Notably, our smallest model, UNO-EELBERT, is just 1.2 MB in size, but achieves a GLUE score within 
4\% 
of that of a standard fully trained model almost 15 times its size.

\bibliography{anthology,custom}

\begin{thebibliography}{30}
\expandafter\ifx\csname natexlab\endcsname\relax\def\natexlab#1{#1}\fi

\bibitem[{Arora et~al.(2018)Arora, Li, Liang, Ma, and
  Risteski}]{arora-etal-2018-linear}
Sanjeev Arora, Yuanzhi Li, Yingyu Liang, Tengyu Ma, and Andrej Risteski. 2018.
\newblock \href {https://doi.org/10.1162/tacl_a_00034} {{Linear Algebraic
  Structure of Word Senses, with Applications to Polysemy}}.
\newblock \emph{{Transactions of the Association for Computational
  Linguistics}}, 6:483--495.

\bibitem[{Chen et~al.(2020)Chen, Frankle, Chang, Liu, Zhang, Wang, and
  Carbin}]{chen2020lottery}
Tianlong Chen, Jonathan Frankle, Shiyu Chang, Sijia Liu, Yang Zhang, Zhangyang
  Wang, and Michael Carbin. 2020.
\newblock {The Lottery Ticket Hypothesis for Pre-trained BERT Networks}.
\newblock \emph{Advances in Neural Information Processing Systems},
  33:15834--15846.

\bibitem[{Devlin et~al.(2019)Devlin, Chang, Lee, and
  Toutanova}]{devlin2018bert}
Jacob Devlin, Ming-Wei Chang, Kenton Lee, and Kristina Toutanova. 2019.
\newblock \href {https://doi.org/10.18653/v1/N19-1423} {{BERT: Pre-training of
  Deep Bidirectional Transformers for Language Understanding}}.
\newblock In \emph{Proceedings of the 2019 Conference of the North {A}merican
  Chapter of the Association for Computational Linguistics: Human Language
  Technologies, Volume 1 (Long and Short Papers)}, pages 4171--4186,
  Minneapolis, Minnesota. Association for Computational Linguistics.

\bibitem[{Ganesh et~al.(2021)Ganesh, Chen, Lou, Khan, Yang, Sajjad, Nakov,
  Chen, and Winslett}]{Ganesh_2021}
Prakhar Ganesh, Yao Chen, Xin Lou, Mohammad~Ali Khan, Yin Yang, Hassan Sajjad,
  Preslav Nakov, Deming Chen, and Marianne Winslett. 2021.
\newblock \href {https://doi.org/10.1162/tacl_a_00413} {{Compressing
  Large-Scale Transformer-Based Models: A Case Study on {BERT} }}.
\newblock \emph{Transactions of the Association for Computational Linguistics},
  9:1061--1080.

\bibitem[{Gokaslan and Cohen(2019)}]{Gokaslan2019OpenWeb}
Aaron Gokaslan and Vanya Cohen. 2019.
\newblock {OpenWebText Corpus}.
\newblock \url{http://Skylion007.github.io/OpenWebTextCorpus}.

\bibitem[{Gordon et~al.(2020)Gordon, Duh, and Andrews}]{gordon2020compressing}
Mitchell Gordon, Kevin Duh, and Nicholas Andrews. 2020.
\newblock {Compressing BERT: Studying the Effects of Weight Pruning on Transfer
  Learning}.
\newblock In \emph{Proceedings of the 5th Workshop on Representation Learning
  for NLP}, pages 143--155.

\bibitem[{Jiao et~al.(2020)Jiao, Yin, Shang, Jiang, Chen, Li, Wang, and
  Liu}]{jiao2020tinybert}
Xiaoqi Jiao, Yichun Yin, Lifeng Shang, Xin Jiang, Xiao Chen, Linlin Li, Fang
  Wang, and Qun Liu. 2020.
\newblock \href {https://doi.org/10.18653/v1/2020.findings-emnlp.372}
  {{TinyBERT: Distilling BERT for Natural Language Understanding}}.
\newblock In \emph{Findings of the Association for Computational Linguistics:
  EMNLP 2020}, pages 4163--4174, Online. Association for Computational
  Linguistics.

\bibitem[{Kaplan et~al.(2020)Kaplan, McCandlish, Henighan, Brown, Chess, Child,
  Gray, Radford, Wu, and Amodei}]{kaplan2020scaling}
Jared Kaplan, Sam McCandlish, Tom Henighan, Tom~B. Brown, Benjamin Chess, Rewon
  Child, Scott Gray, Alec Radford, Jeffrey Wu, and Dario Amodei. 2020.
\newblock \href {http://arxiv.org/abs/2001.08361} {{Scaling Laws for Neural
  Language Models}}.
\newblock \emph{arXiv}, abs/2001.08361.

\bibitem[{Kim et~al.(2020)Kim, Kim, and Lee}]{kim-etal-2020-adaptive}
Yeachan Kim, Kang-Min Kim, and SangKeun Lee. 2020.
\newblock \href {https://doi.org/10.18653/v1/2020.acl-main.364} {{Adaptive
  Compression of Word Embeddings}}.
\newblock In \emph{Proceedings of the 58th Annual Meeting of the Association
  for Computational Linguistics}, pages 3950--3959, Online. Association for
  Computational Linguistics.

\bibitem[{Lam(2018)}]{lam2018word2bits}
Maximilian Lam. 2018.
\newblock \href {http://arxiv.org/abs/1803.05651} {{Word2Bits - Quantized Word
  Vectors}}.
\newblock \emph{arXiv}, abs/1803.05651.

\bibitem[{Lan et~al.(2020)Lan, Chen, Goodman, Gimpel, Sharma, and
  Soricut}]{lan2020albert}
Zhenzhong Lan, Mingda Chen, Sebastian Goodman, Kevin Gimpel, Piyush Sharma, and
  Radu Soricut. 2020.
\newblock {ALBERT: A Lite BERT for Self-supervised Learning of Language
  Representations}.
\newblock In \emph{{Proceedings of the Eighth International Conference on
  Learning Representations}}.

\bibitem[{Lebret and Collobert(2014)}]{lebret-collobert-2014-word}
R{\'e}mi Lebret and Ronan Collobert. 2014.
\newblock \href {https://doi.org/10.3115/v1/E14-1051} {{Word Embeddings through
  Hellinger PCA}}.
\newblock In \emph{Proceedings of the 14th Conference of the {E}uropean Chapter
  of the Association for Computational Linguistics}, pages 482--490,
  Gothenburg, Sweden. Association for Computational Linguistics.

\bibitem[{Liu et~al.(2019)Liu, Ott, Goyal, Du, Joshi, Chen, Levy, Lewis,
  Zettlemoyer, and Stoyanov}]{liu2019roberta}
Yinhan Liu, Myle Ott, Naman Goyal, Jingfei Du, Mandar Joshi, Danqi Chen, Omer
  Levy, Mike Lewis, Luke Zettlemoyer, and Veselin Stoyanov. 2019.
\newblock \href {http://arxiv.org/abs/1907.11692} {{RoBERTa: A Robustly
  Optimized BERT Pretraining Approach}}.
\newblock \emph{arXiv}, abs/1907.11692.

\bibitem[{Norvig(2012)}]{Norvig2012}
Peter Norvig. 2012.
\newblock {English Letter Frequency Counts: Mayzner Revisited or ETAOIN
  SRHLDCU}.
\newblock \url{http://norvig.com/mayzner.html}.
\newblock [Online; accessed 23-October-2023].

\bibitem[{Radford et~al.(2019)Radford, Wu, Child, Luan, Amodei, Sutskever
  et~al.}]{radford2019language}
Alec Radford, Jeffrey Wu, Rewon Child, David Luan, Dario Amodei, Ilya
  Sutskever, et~al. 2019.
\newblock {Language Models are Unsupervised Multitask Learners}.
\newblock \emph{OpenAI Blog}, 1(8):9.

\bibitem[{Raffel et~al.(2020)Raffel, Shazeer, Roberts, Lee, Narang, Matena,
  Zhou, Li, and Liu}]{raffel2020exploring}
Colin Raffel, Noam Shazeer, Adam Roberts, Katherine Lee, Sharan Narang, Michael
  Matena, Yanqi Zhou, Wei Li, and Peter~J Liu. 2020.
\newblock {Exploring the Limits of Transfer Learning with a Unified
  Text-to-Text Transformer}.
\newblock \emph{The Journal of Machine Learning Research}, 21(1):5485--5551.

\bibitem[{Ravi(2017)}]{ravi2017projectionnet}
Sujith Ravi. 2017.
\newblock \href {http://arxiv.org/abs/1708.00630} {{ProjectionNet: Learning
  Efficient On-Device Deep Networks Using Neural Projections}}.
\newblock \emph{arXiv}, abs/1708.00630.

\bibitem[{Ravi and Kozareva(2018{\natexlab{a}})}]{ravi2018self}
Sujith Ravi and Zornitsa Kozareva. 2018{\natexlab{a}}.
\newblock Self-governing neural networks for on-device short text
  classification.
\newblock In \emph{Proceedings of the 2018 Conference on Empirical Methods in
  Natural Language Processing}, pages 887--893.

\bibitem[{Ravi and Kozareva(2018{\natexlab{b}})}]{ravi-kozareva-2018-self}
Sujith Ravi and Zornitsa Kozareva. 2018{\natexlab{b}}.
\newblock \href {https://doi.org/10.18653/v1/D18-1092} {{Self-Governing Neural
  Networks for On-Device Short Text Classification}}.
\newblock In \emph{Proceedings of the 2018 Conference on Empirical Methods in
  Natural Language Processing}, pages 804--810, Brussels, Belgium. Association
  for Computational Linguistics.

\bibitem[{Ravi and Kozareva(2021)}]{ravi-kozareva-2021-soda}
Sujith Ravi and Zornitsa Kozareva. 2021.
\newblock \href {https://aclanthology.org/2021.sigdial-1.7} {{SoDA: On-device
  Conversational Slot Extraction}}.
\newblock In \emph{Proceedings of the 22nd Annual Meeting of the Special
  Interest Group on Discourse and Dialogue}, pages 56--65, Singapore and
  Online. Association for Computational Linguistics.

\bibitem[{Sanh et~al.(2020)Sanh, Debut, Chaumond, and
  Wolf}]{sanh2020distilbert}
Victor Sanh, Lysandre Debut, Julien Chaumond, and Thomas Wolf. 2020.
\newblock \href {http://arxiv.org/abs/1910.01108} {{DistilBERT, a distilled
  version of BERT: smaller, faster, cheaper and lighter}}.
\newblock \emph{arXiv}, abs/1910.01108.

\bibitem[{Sankar et~al.(2021)Sankar, Ravi, and Kozareva}]{sankar2021proformer}
Chinnadhurai Sankar, Sujith Ravi, and Zornitsa Kozareva. 2021.
\newblock {ProFormer: Towards On-Device LSH Projection Based Transformers}.
\newblock In \emph{Proceedings of the 16th Conference of the European Chapter
  of the Association for Computational Linguistics: Main Volume}, pages
  2823--2828.

\bibitem[{Sun et~al.(2019)Sun, Cheng, Gan, and Liu}]{sun2019patient}
Siqi Sun, Yu~Cheng, Zhe Gan, and Jingjing Liu. 2019.
\newblock \href {http://arxiv.org/abs/1908.09355} {{Patient Knowledge
  Distillation for BERT Model Compression}}.
\newblock \emph{arXiv}, abs/1908.09355.

\bibitem[{Tito~Svenstrup et~al.(2017)Tito~Svenstrup, Hansen, and
  Winther}]{tito2017hash}
Dan Tito~Svenstrup, Jonas Hansen, and Ole Winther. 2017.
\newblock {Hash Embeddings for Efficient Word Representations}.
\newblock \emph{Advances in Neural Information Processing Systems},
  30:4935--4943.

\bibitem[{Turc et~al.(2019)Turc, Chang, Lee, and Toutanova}]{turc2019}
Iulia Turc, Ming-Wei Chang, Kenton Lee, and Kristina Toutanova. 2019.
\newblock \href {http://arxiv.org/abs/1908.08962} {{Well-Read Students Learn
  Better: On the Importance of Pre-training Compact Models}}.
\newblock \emph{arXiv}, abs/1908.08962.

\bibitem[{Wang et~al.(2018)Wang, Singh, Michael, Hill, Levy, and
  Bowman}]{wang2019glue}
Alex Wang, Amanpreet Singh, Julian Michael, Felix Hill, Omer Levy, and Samuel
  Bowman. 2018.
\newblock {GLUE: A Multi-Task Benchmark and Analysis Platform for Natural
  Language Understanding}.
\newblock In \emph{Proceedings of the 2018 EMNLP Workshop BlackboxNLP:
  Analyzing and Interpreting Neural Networks for NLP}, pages 353--355.

\bibitem[{{Wikipedia contributors}(2023)}]{enwiki:1168768744}
{Wikipedia contributors}. 2023.
\newblock {Rolling hash --- {Wikipedia}{,} The Free Encyclopedia}.
\newblock
  \url{https://en.wikipedia.org/w/index.php?title=Rolling_hash&oldid=1168768744}.
\newblock [Online; accessed 23-October-2023].

\bibitem[{Wolf et~al.(2019)Wolf, Debut, Sanh, Chaumond, Delangue, Moi, Cistac,
  Rault, Louf, Funtowicz et~al.}]{wolf2020huggingfaces}
Thomas Wolf, Lysandre Debut, Victor Sanh, Julien Chaumond, Clement Delangue,
  Anthony Moi, Pierric Cistac, Tim Rault, R{\'e}mi Louf, Morgan Funtowicz,
  et~al. 2019.
\newblock {Huggingface's Transformers: State-of-the-Art Natural Language
  Processing}.
\newblock \emph{arXiv preprint arXiv:1910.03771}.

\bibitem[{Xue et~al.(2021)Xue, Constant, Roberts, Kale, Al-Rfou, Siddhant,
  Barua, and Raffel}]{xue2021mt5}
Linting Xue, Noah Constant, Adam Roberts, Mihir Kale, Rami Al-Rfou, Aditya
  Siddhant, Aditya Barua, and Colin Raffel. 2021.
\newblock {mT5: A Massively Multilingual Pre-trained Text-to-Text Transformer}.
\newblock In \emph{Proceedings of the 2021 Conference of the North American
  Chapter of the Association for Computational Linguistics: Human Language
  Technologies}, pages 483--498.

\bibitem[{Zafrir et~al.(2019)Zafrir, Boudoukh, Izsak, and
  Wasserblat}]{Zafrir2019quantization}
Ofir Zafrir, Guy Boudoukh, Peter Izsak, and Moshe Wasserblat. 2019.
\newblock \href {https://doi.org/10.1109/emc2-nips53020.2019.00016} {{Q8BERT:
  Quantized 8Bit BERT}}.
\newblock In \emph{2019 Fifth Workshop on Energy Efficient Machine Learning and
  Cognitive Computing - {NeurIPS} Edition ({EMC}2-{NIPS})}, pages 36--39,
  Vancouver, Canada. {IEEE}.

\end{thebibliography}
\bibliographystyle{acl_natbib}

\end{document}